\documentclass{article} % For LaTeX2e
\usepackage{iclr2018_conference,times}
\usepackage{url}
\usepackage{times}  %Required
\usepackage{helvet}  %Required
\usepackage{courier}  %Required
\usepackage{url}  %Required
\usepackage{graphicx}  %Required
\usepackage{subfig}
\usepackage{booktabs}
\usepackage{wrapfig}

\usepackage[pagebackref=true,breaklinks=true,letterpaper=true,colorlinks,bookmarks=false]{hyperref}

\title{Human-like Clustering with Deep Convolutional Neural Networks}

\author{
Ali Borji  \\
Department of Computer Science \\
University of Central Florida, Orlando, FL \\
\texttt{aborji@crcv.ucf.edu}\\
\And
Aysegul Dundar \thanks{Authors contributed equally.}\\
Purdue University \\
West Lafayette, IN \\
\texttt{adundar@purdue.edu}}

\iclrfinalcopy % Uncomment for camera-ready version, but NOT for submission.

\begin{document}

\maketitle

\begin{abstract}
Classification and clustering have been studied separately in machine learning and computer vision. Inspired by the recent success of deep learning models in solving various vision problems (e.g., object recognition, semantic segmentation) and the fact that humans serve as the gold standard in assessing clustering algorithms, here, we advocate for a unified treatment of the two problems and suggest that hierarchical frameworks that progressively build complex patterns on top of the simpler ones (e.g., convolutional neural networks) offer a promising solution. We do not dwell much on the learning mechanisms in these frameworks as they are still a matter of debate, with respect to biological constraints. Instead, we emphasize on the compositionality of the real world structures and objects. In particular, we show that CNNs, trained end to end using back propagation with noisy labels, are able to cluster data points belonging to several overlapping shapes, and do so much better than the state of the art algorithms. The main takeaway lesson from our study is that mechanisms of human vision, particularly the hierarchal organization of the visual ventral stream should be taken into account in clustering algorithms (e.g., for learning representations in an unsupervised manner or with minimum supervision) to reach human level clustering performance. This, by no means, suggests that other methods do not hold merits. For example, methods relying on pairwise affinities (e.g., spectral clustering) have been very successful in many scenarios but still fail in some cases (e.g., overlapping clusters).
\end{abstract}

\section{Introduction}
Clustering, a.k.a unsupervised classification or nonparametric density estimation, is central to many data-driven domains and has been studied heavily in the past. The task in clustering is to group a given collection of unlabeled patterns into meaningful clusters such that objects within a cluster are more similar to each other than they are to objects in other clusters. Clustering provides a summary representation of data at a coarse level and is used widely in many disciplines (e.g., computer version, bioinformatics, text processing) for exploratory data analysis (a.k.a pattern mining) as well as representation learning (e.g., bag of words). Despite the introduction of thousands of clustering algorithms in the past~\cite{aggarwal2013data}, some challenges still remain. For instance, existing algorithms fall short in dealing with different cluster shapes, high dimensions, automatically determining the number of clusters or other parameters, large amounts of data, choosing the appropriate similarity measure, incorporating domain knowledge, and cluster evaluation. Further, no clustering algorithm can consistently win over other algorithms, handle all test cases, and perform at the level of humans. 

Deep neural networks have become a dominant approach to solve various tasks across many fields. They have been proven successful in several domains including computer vision~\cite{krizhevsky2012imagenet}, natural language processing~\cite{collobert2011natural}, and speech recognition~\cite{dahl2012context} for tasks such as scene and object classification~\cite{krizhevsky2012imagenet}, pixel-level labeling for image segmentation~\cite{long2015fully,zheng2015conditional}, modeling attention ~\cite{borji2013state,borji2013quantitative}, image generation~\cite{goodfellow2014generative}, robot arm control~\cite{levine2015end}, speech recognition~\cite{graves2014towards}, playing Atari games~\cite{mnih2015human} and beating the Go champion. %~\cite{silver2016mastering}. 

%All of the above techniques largely avoid sophisticated pipeline implementations and human-in-the-loop tuning by adopting the concept of training the networks to learn the target problem directly. 

Deep Convolutional Neural Networks (CNNs)~\cite{lecun1998gradient} have been particularly successful over vision problems. One reason is that nearby pixels in natural scenes are highly correlated. Further natural objects are compositional. These facts allow applying the same filters across spatial locations (and hence share weights), and build complex filters from simpler ones to detect high-level patterns (e.g., object parts, objects). We advocate that these properties are highly appealing when dealing with clustering problems. For instance, the classic two half moons example can be solved by applying a filter that is selective to each half moon. Or, when two clusters with different shapes overlap, the problem can be solved by having filters responding to each shape. Solving these cases is very challenging by just looking at local regions around points and being blind to the high-level patterns. Incorporating domain knowledge, while working in some cases, does not give a general solution for solving all clustering problems. The human visual system easily solves these 2D problems because it is a general system with a rich set of learned or evolved filters. We believe that deep CNNs, although imperfect models of the human vision as they lack feedback and lateral connections carry a huge promise for solving clustering tasks. Further, as we will argue, they offer a unified solution to both classification and clustering tasks.

The current demarcation between classification and clustering becomes murky when we notice that researchers often refer to human judgments in evaluating the outcomes of clustering algorithms. Indeed, humans learn quite a lot about the visual world during their life time. Moreover, the structure of the visual system has been fine-tuned through the evolution. Thus, certainly, there is a learning component involved which has been often neglected in formulating clustering algorithms. While this is sensible from an application point of view (e.g., pattern mining), not only it limits the pursuit for stronger algorithms but also narrows our understanding of human vision.

Learning techniques have been utilized for clustering in the past (e.g.,~\cite{bach2004learning,pinheiro2016learning}), for example for tuning parameters (e.g.,~\cite{bach2004learning}). Deep networks have also been exploited for clustering (e.g.,~\cite{hsu2015neural,hershey2016deep,wang2016learning}). However, to our knowledge, while CNNs have been already adopted for image segmentation, so far they have not been exploited for generic clustering. Our goal is to investigate such possibility. To this end, instead of borrowing from clustering to do image segmentation, we follow the opposite direction and propose a deep learning based approach to clustering.

Our method builds on the fully convolutional network literature, in particular, recent work on edge detection and semantic segmentation which utilize multi-scale local and non-local cues~\cite{ronneberger2015u}. Thanks to a high volume of labeled data, high capacity of deep networks, powerful optimization algorithms, and high computational power, deep models win on these tasks. We are also strongly inspired by the works showing the high resemblance between human vision mechanisms and CNNs from behavioral, electrophysiological, and computational aspects (e.g.,~\cite{yamins2014performance,dicarlo2007untangling,lecun1998gradient,krizhevsky2012imagenet,borji2014human}. Our study enriches our understanding of the concept of clustering and its relation to classification.

\section{Related Work}
\label{sec:relatedwork}
\noindent \textbf{Data clustering algorithms.} Clustering approaches can be broadly classified into hierarchical and partitional types~\cite{aggarwal2013data}. Hierarchical approaches either start with a single big cluster and recursively split it into smaller ones (top-down) or begin with several small clusters, and then gradually merge them (bottom-up or agglomerative).
A classic example of partitional methods is the k-means~\cite{lloyd1982least} algorithm which minimizes the sum of squared errors between data points and their nearest cluster centers. Some frequently used algorithms include spectral clustering~\cite{chung1997spectral,ng2001spectral}, expectation maximization (EM)~\cite{dempster1977maximum}, Meanshift~\cite{comaniciu2002mean}, and non-negative matrix factorization (NMF)~\cite{lee2001algorithms}. Density-based algorithms, a class of partitional methods, classify points by identifying regions heavily populated with data. Some examples include DBSCAN~\cite{ester1996density}, GDBSCAN~\cite{sander1998density}, OPTICS~\cite{ankerst1999optics}, and CFSFDP~\cite{rodriguez2014clustering}. Neural networks have also been utilized to build clustering algorithms (e.g., ARTMAP~\cite{carpenter1991artmap} and SOM~\cite{kohonen1998self}).

\noindent \textbf{Boundary assignment.} 
This problem has also been studied under the names of boundary completion, border ownership, occlusion handling, figure-ground interpretation or segmentation in cognitive sciences, computational neuroscience (e.g.,~\cite{zhou2000coding}) and computer vision. It has long been recognized as an important human ability for scene understanding and perception. Here, the main goal is to tell which region a border belongs to or to identify an object in the image and separate it from the background. The Gestalt psychologists were among the first to study these processes (e.g.,~\cite{koffka2013principles}). They pointed out that visual perception tends to assign dividing edges to objects, not to the background (e.g., Kanizsa triangle). Research in computer vision on detecting occlusion relations in natural images was stimulated by the construction of the BSDS border ownership dataset~\cite{ren2006figure}. Deep learning methods have recently been exploited for solving the boundary assignment problem (e.g.,~\cite{wang2015doc}).

\noindent \textbf{Semantic segmentation.}
%super pixels in computer vision
The goal in these techniques is to assign structured semantic labels — typically, object class labels — to individual pixels in images. This problem has been studied extensively over decades, yet remains challenging. This is mainly because objects can appear widely differently in images due to significant variations is the pose, scale, lighting, occlusion, background clutter, etc. However, in spite of such challenges, the techniques based on deep learning demonstrate impressive performance in the standard benchmark datasets such as PASCAL VOC~\cite{everingham2010pascal} or MS COCO~\cite{lin2014microsoft}. Most deep learning based approaches pose semantic segmentation as a pixel-wise classification problem (e.g.,~\cite{long2015fully}). Although these approaches have achieved good performance compared to previous methods, they demand a large number of segmentation ground-truths. 
Clustering scenarios we are studying in this paper are quite different than image segmentation and instance segmentation challenges. In clustering, the input is sparse and occlusion is see through, therefore a combination of detection and segmentation can not give the instance clusters.

\noindent \textbf{Deep learning for clustering.}
Currently, there is only a little work on exploiting deep neural network for clustering. Hsu and Kira~\cite{hsu2015neural} utilized pairwise constraints to train a neural network to perform clustering. In~\cite{chen2015deep}, deep belief networks~\cite{hinton2009deep} were used for non-parametric clustering. Wang et al.~\cite{wang2016learning} proposed a task-specific deep architecture for clustering based on sparse coding. Some works have focused on learning representations (or embeddings) for clustering. Here the idea is to first transform the original data to a lower dimensional space with a nonlinear mapping $f_\theta : X \rightarrow Z$, where $\theta$ are parameters and Z is the latent feature space. Adopting this, Xie et al.~\cite{xie2016unsupervised} proposed an approach for image clustering. Song et al.~\cite{song2013auto} defined a new objective function by considering the reconstruction error from an auto-encoder network and restricting the distance in the learned space between data and their corresponding cluster centers. Hershey et al.~\cite{hershey2016deep} proposed a deep clustering approach to solving acoustic source separation. Rather than directly estimating signals or masking functions, they trained a deep network to produce spectrogram embeddings.~\cite{chen2015deep} explored the possibility of employing deep learning in graph clustering. They first learned a nonlinear embedding of the original graph by an autoencoder, followed by k-means algorithm on the embedding to obtain the final clustering result.

\section{Model Description}
\label{sec:model}

\begin{figure}[t]
	\centering
	%\footnotesize
    %\vspace{-10pt}
    \includegraphics[width=\linewidth]{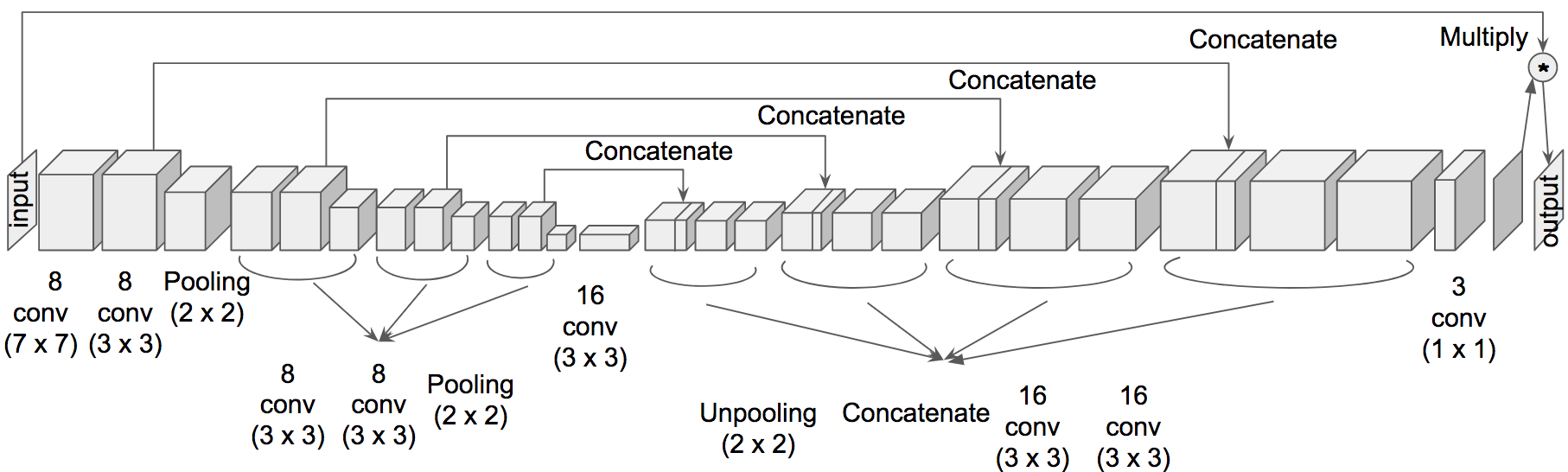} \\
    \vspace{-5pt}
	\caption{\small{U-Net architecture~\cite{ronneberger2015u} adopted in this work.}}
	\label{fig:arch}
    \vspace{-18pt}
\end{figure}

We are motivated by three observations. First, CNNs have been very successful over a variety of vision problems such as semantic segmentation, edge detection, and recognition. Second, CNNs learn representations through several stages of non-linear processing, akin to how the cortex adapts to represent the visual world~\cite{yamins2014performance}. Similar to other biological models of visual cortex (e.g., HMAX~\cite{serre2007robust}), these models capture aspects of the organization of the visual ventral stream. Third, clustering methods are often evaluated against human perception motivating biologically inspired solutions. 

Our strategy parallels recent work in using CNNs for semantic segmentation. A crucial difference, however, is that cluster identities (class labels) are not important here. For instance, if a triangle and a circle exist in the image, the shape labels can be anything as long as clusters are correctly separated. Unlike previous work, instead of learning embeddings, we use back propagation via stochastic gradient descent to optimize a clustering objective to learn the mapping, which is parameterized by a deep neural network. In this way, there is no need to specify parameters like number of clusters, distance measure, scale, cluster centers, etc.

\subsection{The network architecture}

Figure~\ref{fig:arch} shows the proposed deep network architecture which is based on the U-Net \cite{ronneberger2015u}; an encoder-decoder with skip connections. The input is a binary image with a single channel, also shown in Figure~\ref{fig:stimuli}. Input is fed to five stacks of [convolution, convolutions, pooling] modules. These are followed by five stacks of decoder modules [convolution, convolution, upsampling]. Skip connections from mirrored layers in the encoder are fed to decoder stacks. Such skip connections recover the spatial image information which might have been lost through successive convolution and pooling operations in the encoder.
Finally, three $1 \times 1$ filters are applied to collapse the convolutional maps to three channels that can cluster three objects (one channel per cluster). Each convolution layer in the decoder module has 16 filters and is followed by a ReLU layer except the last convolution layer which uses Sigmoid activation. Then, the pointwise multiplication of the output and the input is calculated to generate the final cluster map (denoted as the output map in Figure~\ref{fig:stimuli}). This multiplication removes the background pixels from the output, giving us only the prediction for points that are of interest.

\subsection{Training procedure and ground truth}

To implement our model, we use the Keras~\cite{chollet2015keras} platform. We use $128 \times 128$ binary images as inputs (see Figure~\ref{fig:stimuli}). The corresponding ground truth map for each input is generated as follows: the points belonging to the topmost cluster are assigned label 0, descending top-down in the image, points belonging to the next cluster are assigned label 1, and so on. This process is repeated until all clusters are labeled. This way the labels are independent of the object shapes. This makes our training scheme different from classification and segmentation methods where each object is always assigned the exact same label. Notice that here we are mainly interested in separating the objects from each other rather than correctly classifying them. Despite this, as we will show later, the network is able to cluster objects with the same shape successfully. We use the mean squared error loss and train the network with the Adam optimizer~\cite{kingma2014adam}. Batch size is set to 16 and learning rate to 0.001.

\section{Experimental Setup}
\label{sec:experiment}

\subsection{Synthetic data}

\begin{figure}[t]
\centering
\subfloat[]{\includegraphics[width = .9\linewidth]{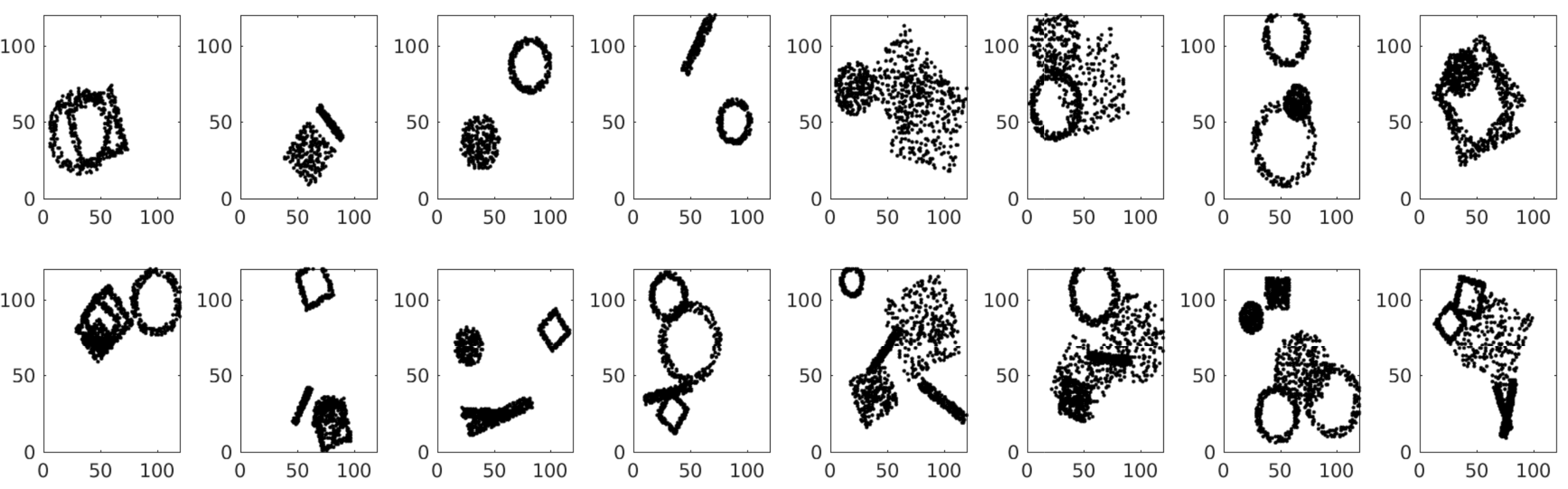}}\\ 
\vspace{-10pt}
\subfloat[]{\includegraphics[width = .9\linewidth]{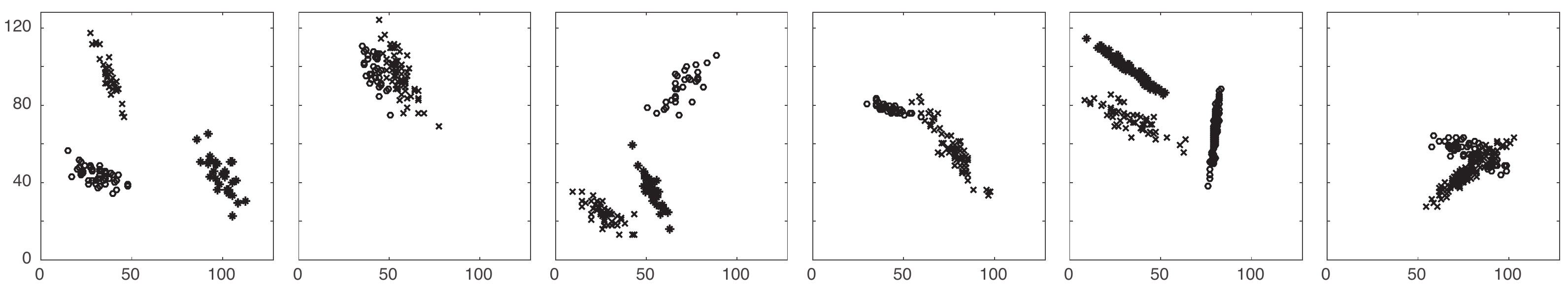}} 
\vspace{-10pt}
\caption{\small{Examples stimuli. Parameters include number of objects, shape type, point density, object scale, rotation, and location. Parameter ranges are set to increase the chance of overlap among objects. (a) Stimuli generated with shapes. Shapes include circle(filled), ring, square(filled), square ring, and bar. (b) Stimuli generated with Gaussian distributions. Gaussian clusters are shown with different markers for the illustration purpose.}}
\label{fig:stimuli}
%\vspace{-18pt}
\end{figure}

%Here, we explain the process of generating the synthetic images to train and test our model.

\noindent \textbf{Geometric shape stimuli.} Objects are parametrized using several variables including $S$ = \{Circle, Ring, Square, SquareRing, Bar\}, $O = \{1,\ldots,m\}$, $D = [200 \ \ 300]$, and $SC = [10 \ \ 30]$. They, in order, denote the set of possible shapes, the set of the possible number of objects to place in the image, the interval of point densities for an object, and the interval of possible object scales. 

To generate an image, first a number $k \in O$, indicating the number of objects is randomly drawn. The following is then repeated $k$ times. Random density and scales are chosen for clusters. The cluster is randomly rotated $\theta$ degrees ($\theta \in [0, 2\pi]$) and shifted such that it remains within the image boundary. The output is then saved as a $128 \times 128$ pixels binary image (1 for shape pixels and 0 for background) to be fed to CNN. Ground truth of clusters is saved for training and evaluation.

\noindent \textbf{Gaussian mixture distribution.} To randomly generate a Gaussian Mixture Density distribution, a 2D mean vector $M = [x \ y]$ is randomly sampled $(x, y \in [20 \ 100 ])$. A random matrix $A = [a \ b; c \ d]$ is generated with elements in $[0 \ 1]$. The matrix $A'A$ which is symmetric and positive semidefinite is chosen as the covariance matrix for a Gaussian. 
The same is repeated to assign random mean and covariances matrices for $m$ Gaussians ($m \in \{2, 3\}$) in the image. After having the Gaussian distributions, we randomly sample $D \in [100 \ 400]$ points from each Gaussian and form the input and output images as in the shapes case.

Some sample generated images are shown in Figure~\ref{fig:stimuli}. While it seems easy for humans to find the clusters in these images, as will be shown in the next section, our stimuli poses a serious challenge to current best clustering algorithms. In particular, current clustering algorithms fail when clusters occlude each other. The reason is that they lack a global understanding of the patterns. 

\begin{figure*}[t]
	\centering
	%\footnotesize
    \includegraphics[width=\linewidth]{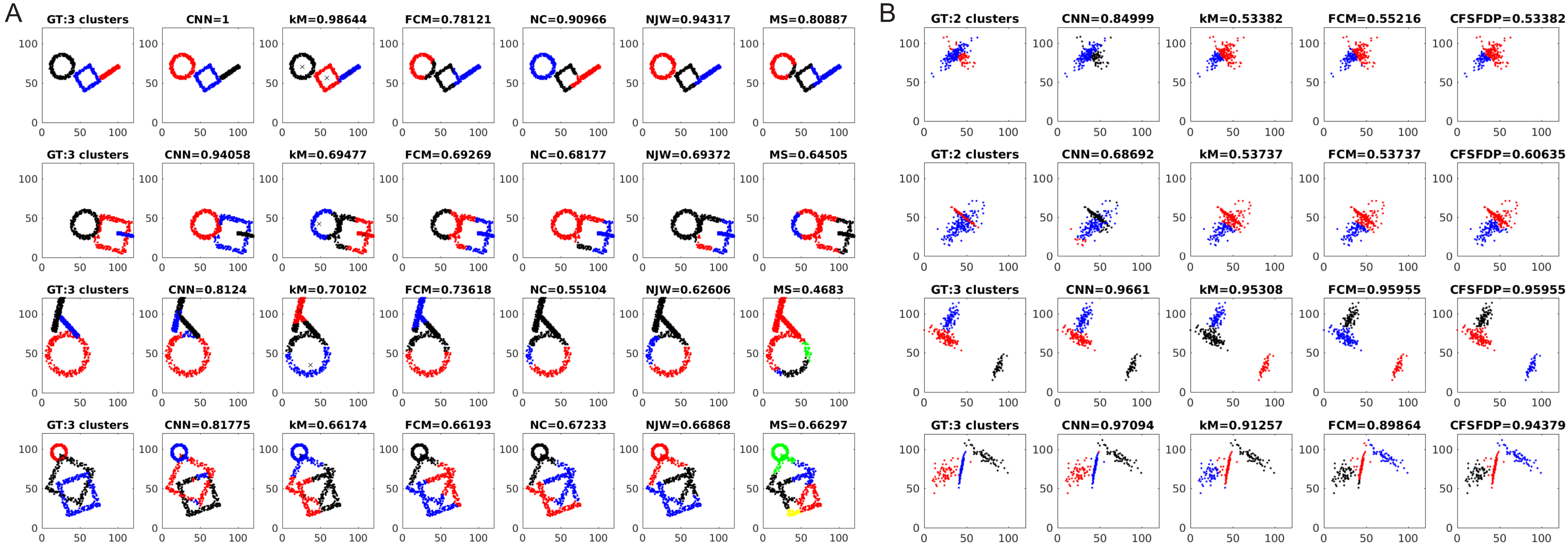} \\
    \vspace{-5pt}
	\caption{\small{Sample images and the output of our clustering algorithm compared to other methods (A) in experiment three (3 clusters, 3 shapes) and (B) in experiment five (2 or 3 clusters, Gaussian mixtures).}}
	\label{fig:exp1}
    \vspace{-15pt}
\end{figure*}

\subsection{Evaluation metric}
\label{evaluation}
Since our purpose is to do clustering, and not classification, prediction score is not sensible to evaluate the performance. Instead, we use rand index which is used to evaluate clustering methods ~\cite{rand1971objective}. The metric is as follows: Given $n$ points in the image, a binary matrix of size $n^2$ is formed where each element indicates whether two points belong to the same cluster or not. A similar matrix is created for the prediction of each model. Notice that the order of points is preserved when constructing these matrices. Then, the Hamming distance between the ground truth matrix and the prediction matrix is calculated which determines the fraction of cases they do not agree with each other (i.e., error rate).
%This is basically equivalent to the rand index used to evaluate clustering methods [].

\subsection{Benchmark algorithms}

We used the following classic and state of the art clustering methods as the benchmark algorithms:\\
\noindent \textbf{k-Means (KM)}~\cite{lloyd1982least}, minimizes the sum of squared errors between data points and their nearest cluster centers. It is a simple and widely used method. 

\noindent \textbf{Fuzzy C-Means (FCM)}~\cite{dunn1973fuzzy} assigns soft labels to data points meaning that each data point can belong to more than one cluster with different degrees of membership.

\noindent \textbf{Spectral Clustering.} We employ two spectral clustering algorithms. These algorithms perform a spectral analysis of the matrix of point-to-point similarities instead of estimating an explicit model of data distribution (as in k-Means). The first one is the Normalized Cut (NC) proposed by Shi and Malik~\cite{shi2000normalized}. Here, first, a graph is built from the image (pixels as nodes, edges as similarities between pixels). Then, algorithm cuts the graph into two subgraphs. The second algorithm, known as Ng-Jordan-Weiss (NJW)~\cite{ng2002spectral} is a simple and significant example of spectral clustering which analysis the eigenvectors of the Laplacian of the similarity matrix. Spectral clustering has been shown to work well on data that is connected but not necessarily compact or clustered within convex boundaries.

\noindent \textbf{Mean shift (MS)}~\cite{comaniciu2002mean} iteratively seeks the modes of a density function from discrete samples of that function. Mean Shift performs as follows. First, it fixes a window around each data point. Then, computes the mean of data within each window. Finally, shifts the window to the mean and repeats till convergence.

\noindent \textbf{Clustering by fast search and find of density peaks (CFSFDP)}~\cite{rodriguez2014clustering} method seeks the modes or peaks of a distribution. It works as follows: 1) For each point, its local density is computed (i.e., number of points in its neighborhood), 2) For each point, its distance to all the points with higher density is computed and the minimum value is sought, 3) 
A plot is made showing the minimum distance for a point as a function of the density. The “outliers” in this plot are the cluster centers, 5) Finally, each point is assigned to the same cluster of its nearest neighbor of higher density. The input to this algorithms is the pairwise distance matrix. To find outliers (i.e., cluster centers), we find the point [max$_x$ \  max$_y$] in this plot and then find $q$ closest points to this point with $q$ being equal to number of ground truth clusters. We use the Gaussian cut off kernel. % in this model.

All algorithms are provided with the actual number of clusters that exist in the input image, except the CNN and MS algorithm which are supposed to automatically determine the number of clusters in the process of clustering. Euclidean distance is used in both k-Means and FCM as the distance measure. All algorithms are optimized for their best performance in each experiment (e.g., by varying the type of affinity, scale, and normalization). The last three algorithms have been very successful for solving the perceptual grouping problem in computer vision.

\subsection{Experiments and Results}
%Num data, train, test, validation splits, activations functions, loss, etc

%visualization of intermediate layers

We run a total of nine experiments. The first six experiments aim to evaluate the performance of the proposed method with respect to the benchmark algorithms and contain different set-ups with geometric shape and Gaussian mixture data. The last three experiments test the robustness of the proposed method. 

\noindent \textbf{Experiment 1:} We generate images with 2 objects randomly chosen from 5 shapes. We use 1800 training and 200 testing images. The goal here is to study the effect of cluster heterogeneity on the results. Results of the experiment 1, the first row of Table~\ref{table:results}, show that CNN is able to successfully cluster the data over the easy cases when two different objects are in the image. Yet, these images challenge the other algorithms.

\noindent \textbf{Experiment 2:} We generate 200 test images with 2 objects of the same shape type (one of the 5 shapes). We use the network trained in our 1st experiment. This experiment concentrates on the proposed method's ability to cluster shapes that look very similar to each other. This case is important since CNNs are known to be very good at generalizing. Accuracies, presented in Table~\ref{table:results}, drop compared to experiment 1 but CNN still outperforms other models.
%Our model sometimes fails when objects touch each other since points in the shared area can belong to any of the objects. 

\noindent \textbf{Experiment 3:} We generate images with 3 objects randomly chosen from 3 shapes (Ring, SquareRing, and Bar). We use 2700 training images.
Figure~\ref{fig:exp1}(A) illustrates the output of models over 4 examples. In the first example, the bar slightly touches the rotated square. While CNN is capable of handling this case, other models bleed around the point of touch. In example two, the square is occluded with the bar. Again, while CNN is able to correctly assign the cluster labels to occluded objects, other models are drastically hindered. Our model scores 81.2\% while other models perform no better than 70\%. Similar patterns can be observed over the other two examples. These findings also hold over images with varying numbers of objects, and parameters (e.g., 5 objects, 5 shapes) as well as Gaussian clusters (see Figure~\ref{fig:exp1}(B)).

\noindent\textbf{Experiment 4:} We consider a more challenging case of having 3 objects sampled from 5 shapes. Here, the train set contains 7000 images.
Results are lower in experiment 4 compared to the first 2 experiments. We find that performance drops as we add more objects or increase the shape variety.

\begin{figure*}[t]
	\centering
	%\footnotesize
    \includegraphics[width=1\linewidth]{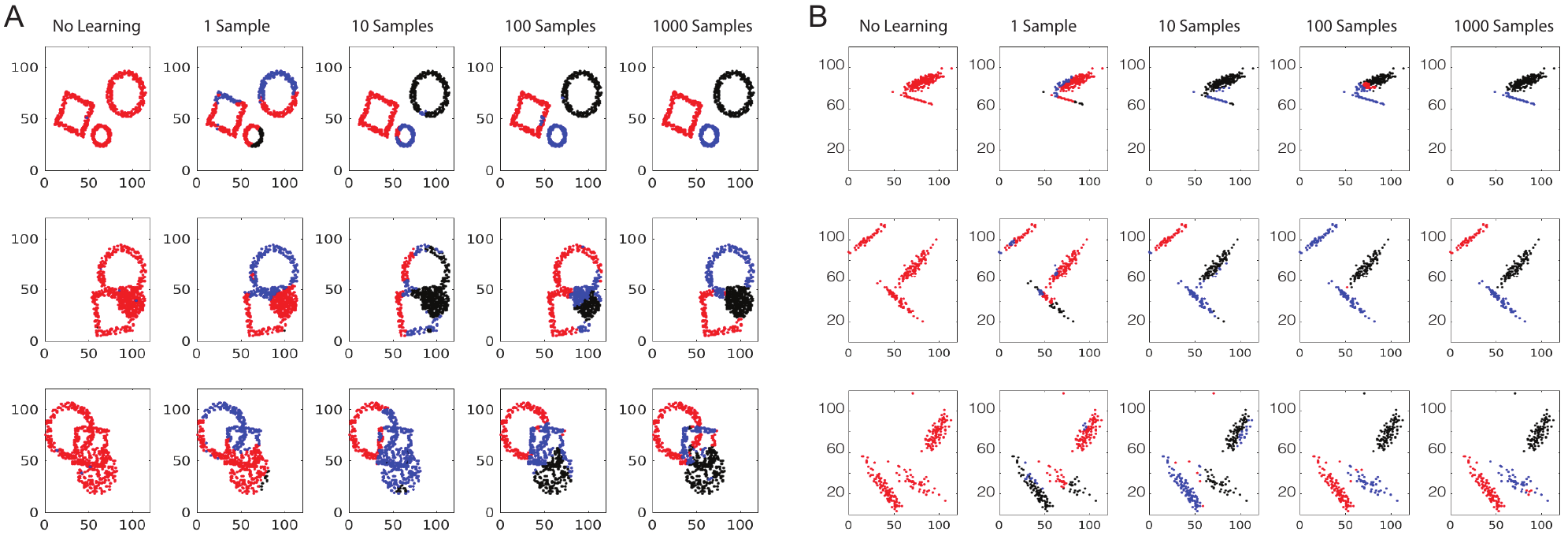} \\
    \vspace{-5pt}
	\caption{\small{Results of experiment 8. Sample outputs of the CNN trained with different number of images (n = 200). A: for shapes and B: for Gaussian distributions.}}
    \vspace{-15pt}
	\label{fig:learning}
\end{figure*}

\begin{wraptable}{r}{7.1cm}
\begin{center}
\vspace{-7pt}
	\setlength\tabcolsep{3pt}
    \caption{\small {Quantitative comparison of clustering algorithms over shape stimuli. The best one is highlighted in \textbf{bold}. Rows in each experiment show means and standard deviations, respectively.}}
    \vspace{-7pt}
    \begin{scriptsize}
	\begin{tabular}{lccccccc} 
Model      & CNN    & kM & FCM  & NJW   &  SC  & MS  & CFSFDP  \\ \specialrule{0.1em}{2pt}{2pt}
Exp 1   & \textbf{0.916}	&0.850	&0.858 &0.816&	0.805	&0.748 & 0.771 \\ 
\hspace{30pt} & 0.131 &	0.173 &	0.165&	0.220	&0.213	&0.181 & 0.207\\ 
Exp 2    & \textbf{0.895} &	0.849 &	0.859	&0.813	&0.811	&0.749 & 0.756   \\ 
\hspace{30pt}   & 0.151 &	0.176	&0.168	&0.224&	0.217&	0.192 & 0.212\\ 
Exp 3    &  \textbf{0.895} &	0.770 &0.771 &0.721	 &  0.760&0.714 & 0.728 \\ 
\hspace{30pt}   	&0.098	&0.107	&0.103	&0.138&	0.139&	0.116 & 0.134\\ 
Exp 4   &   \textbf{0.871}	& 0.798	& 0.804	& 0.703	&0.739 &	0.714 & 0.729 \\ 
\hspace{30pt}   & 0.117 &	0.120	& 0.117	& 0.190	& 0.153	& 0.167  & 0.130 \\
	\end{tabular}
    \end{scriptsize}
	\label{table:results}
    \end{center}
    \vspace{-18pt}
\end{wraptable}

\noindent \textbf{Experiments 5 \& 6:} We use images that can have 2 or 3 different Gaussian clusters both in training and testing data. We do not give the number of clusters to CNN and MS algorithms during testing. Experiment 5 has clusters that have (100-400) number of points, whereas experiment 6 has denser clusters (400-700) points. As it can be seen in Table~\ref{table:resultsGMM}, similar to shapes, CNN is superior to other approaches in clustering Gaussian distributions. 

\noindent \textbf{Experiment 7:} We further investigate the behavior of CNN to different number of clusters in images. Applying a network trained over 3 objects to cluster 2 objects (over shapes), shows a drop in performance compared to the situation where the network was trained on 2 objects (from 91.6\% to 88\%). The accuracy is still high and better than other algorithms. This result suggests that CNN model is not fitted to a certain task and learns about what constitutes a cluster.

\begin{wraptable}{r}{7.6cm}
\begin{center}
\vspace{-7pt}
\small{
	\setlength\tabcolsep{4pt}
    \caption{\small{Quantitative comparison of algorithms over Gaussian stimuli. CNN is over 15K training data and 1K test.}} %First row is with 100-400 density while second row is 400-700.}
    \vspace{-7pt}
    \begin{scriptsize}
\begin{tabular}{lccccccc} 
	Model      & CNN    & kM & FCM  & NJW   &  SC  & MS  & CFSFDP \\ \specialrule{0.1em}{2pt}{2pt}
Exp 5  & 
\textbf{0.920} &  0.868  &  0.870  &  0.844 &   0.853 &   0.838  & 0.878   \\
\textit{\underline{d(100-400)}}		& 0.112  &   0.138 &   0.135 &   0.158 &   0.145  &  0.176  & 0.152 \\
Exp 6   & 
\textbf{0.916} &  0.859  &  0.854  &  0.845 &  0.851 &  0.840  & 0.895   \\
\textit{\underline{d(400-700)}}		& 0.116  & 0.136 & 0.135 & 0.150 &   0.138  &  0.180  & 0.143 \\
	\end{tabular}
    \end{scriptsize}
	%\vspace{1pt}
	\label{table:resultsGMM}
   % \vspace{-20pt}
   \vspace{-15pt}
    }
    \end{center}
\end{wraptable}

\begin{wrapfigure}{R}{0.6\textwidth}
\centering
\vspace{-55pt}
  \subfloat[]{\includegraphics[width = 1\linewidth]{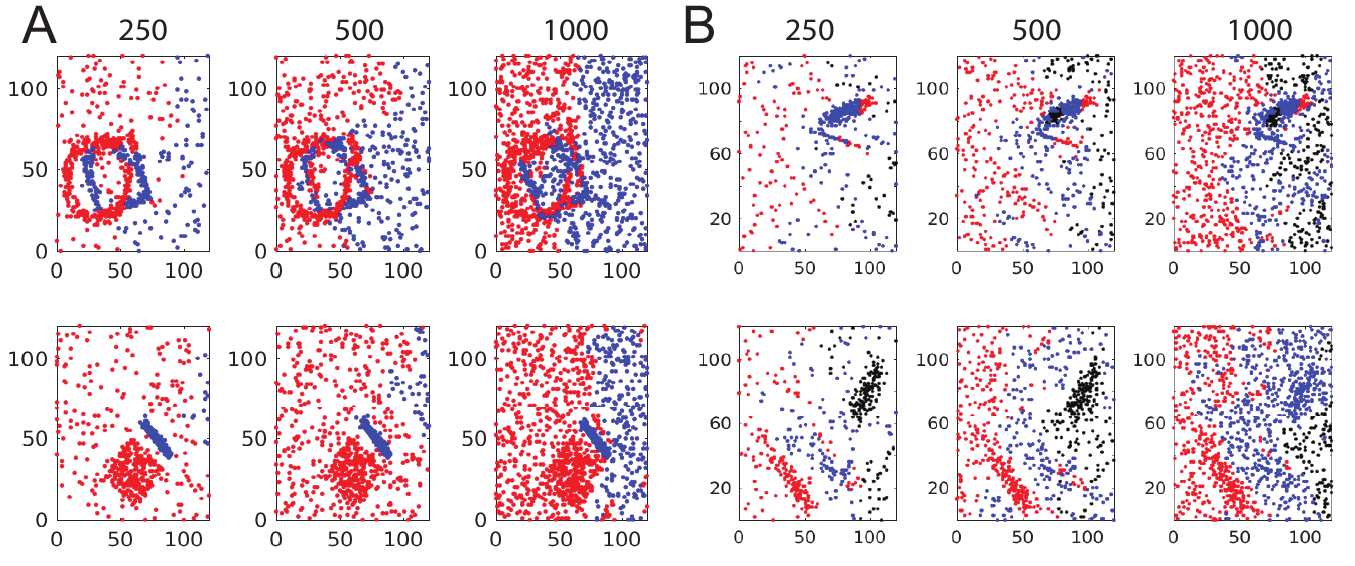}\label{fig:noise}}\\
      \vspace{-10pt}
\subfloat[]{\includegraphics[width = 1\linewidth]{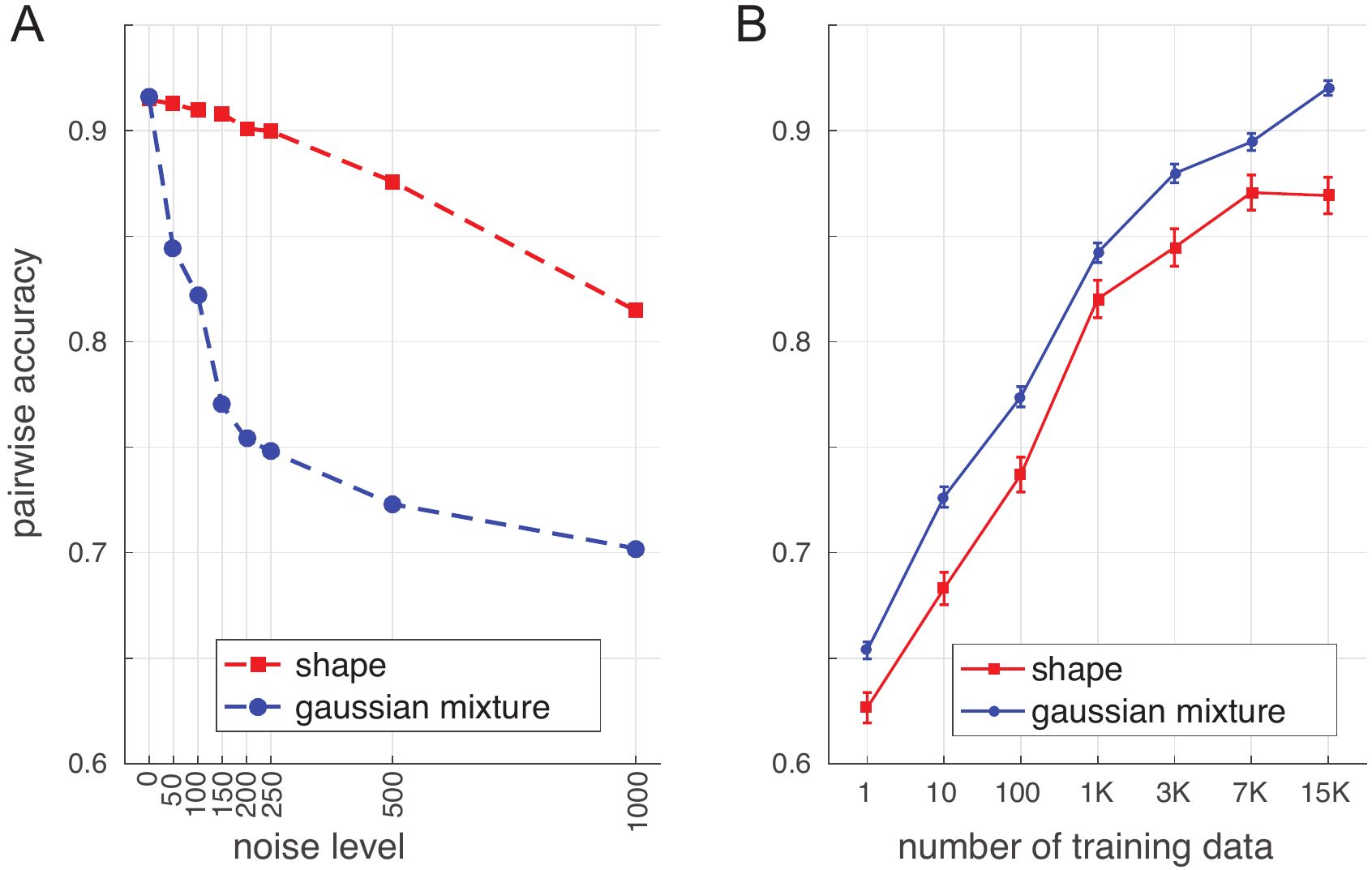}\label{fig:noise_learning_graph}}  
 	    %\vspace{-5pt}
    \caption{\small{(a) Results of experiment 9. Sample outputs of CNN, trained over noiseless data, over 3 images corrupted with \{250, 500, 1000\} amount of noise. A: Shape clusters, B: Gaussian clusters. (b) A: Results of experiment 9, noise level versus pairwise accuracy. B: Results of experiment 8, number of training data samples versus pairwise accuracy. }}
    \vspace{-15pt}
\end{wrapfigure}

\noindent \textbf{Experiment 8:} Here, we investigates the generalization power using different amounts of train data for shapes and Gaussian distributions. A model is first trained over a variable number of $n$ images ($n \in \{1, 10, 100, 1000, 3000, 7000\}$) and is then tested over 200 test images. To measure the lower bound, we also test a randomly initialized network that has not seen any training sample. Results are shown in Figure~\ref{fig:learning}. The left and right panels in this figure show models' predictions for shapes and Gaussian clusters respectively. Expectedly, it shows that the model fails without training (i.e., model with its weights randomly initialized). Here, the model achieves 34\% accuracy. To our surprise, training with only a single sample leads to a better than chance performance of 62.7\% (chance is 50\%) for shapes. Training with only 10 samples gives a descent accuracy (68.3\%). Increasing the training size to 100 leads to a performance comparable to some algorithms (NJW). With 1000 examples, the model is as good as k-Means and FCM and better than other methods. This result is illustrated in Figure~\ref{fig:noise_learning_graph}.B. Results on shape and Gaussian clusters follow a similar trend. In sum, results of this experiment suggest that our model is quite efficient in learning from a few samples and generalizes well to unseen cases.

\noindent \textbf{Experiment 9:} We test the robustness of our model to noise by randomly switch some points in the background to 1 (i.e., making them shape points). Figure~\ref{fig:noise} shows some examples. The pre-trained models from the first and fifth experiments, trained over noiseless data, are applied to the noisy test images. We do not set a threshold in the output so the noises also receive a cluster id. In this experiment, we are interested in analyzing the effect of injecting additional noise on the real clusters. To measure the performance over noisy images, we discard the noise pixels in the evaluation. 
Two inputs, corrupted with 3 levels of noise, are shown in Figure~\ref{fig:noise} for both shape and Gaussian clusters. Figure~\ref{fig:noise_learning_graph}.A shows a gradual decrease in performance by increasing the amount of noise. Network is affected by noise more when the clusters are Gaussians since it is a harder task as can be seen from Figure~\ref{fig:noise}. Even with highly degraded images especially for shapes, the model does reasonably well. Therefore, our model, unlike other methods, is robust to noise. 

We repeated experiments 4 and 5 with randomly labeled clusters as opposed to our ground-truth generation which follows a top-down ordering. The networks did not learn from such ground truth and give accuracy of 54\% for experiment 4, and 63.62\% for experiment 5 (still above chance). 

%Table~\ref{table:results} shows the results of the first five experiments. CNN wins over all models in all experiments with a large margin (e.g., about 7\% improvement over the second best in experiment 1). This large margin hints that maybe even the best optimization of the compared algorithms will not be able to compete with the proposed CNN. All other models perform about the same. 

\subsection{User study}
To answer whether the clustering results by our method are closer to human perception of clusters, we ran a user study by asking subjects to choose the output of the model, among three alternatives, that best describes a set of data points. Three methods include CNN, FCM, and CFSFDP, which were the three best in the Gaussian experiment.

\noindent \textbf{Subjects:} Fourteen students from the University of [masked] (8 women, mean age 24.6) participated. They all had normal or corrected-to-normal vision. 

\begin{wrapfigure}{R}{0.45\textwidth}
\centering
\vspace{-55pt}
\includegraphics[width=0.5\textwidth]{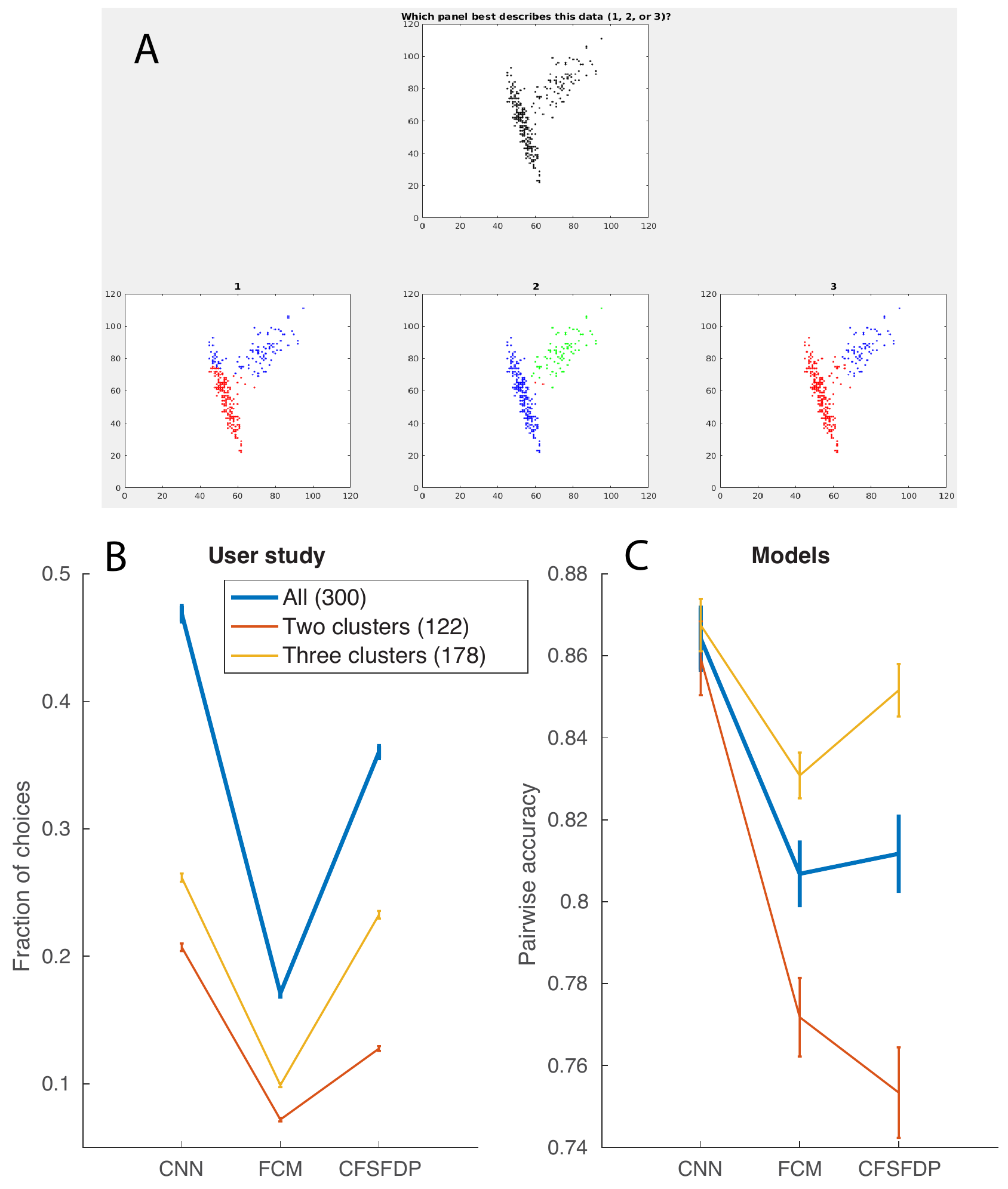}
    \vspace{-10pt}
	\caption{\small{A) A sample trial in the user study, B) Fraction of the cases where subjects chose models and break downs over clusters (all, 2, or 3), and C) Performance of the models over the stimuli.}}
	\label{fig:user}
    \vspace{-15pt}
\end{wrapfigure}

\noindent \textbf{Procedure:} Fig.~\ref{fig:user}.A shows a sample trial in which an image was shown at the top and 3 alternative solutions were shown at the bottom. The locations were chosen randomly for each algorithm and were counter balanced over all stimuli. Subjects sat 80 cm away from a 17" LCD monitor. They were instructed to select the clustering that they thought best describes the data. After pressing a key (1, 2, or 3 identifying one of the algorithms), screen moved to the next trial and so on. The experiment took about 40 minutes to complete. There was no time constraint for each trial. There was a practice session of five trials for the subjects to get familiar with the goal of the experiment. 

%An example trial is shown in Fig.~\ref{fig:exp1}.A. Three methods include ours, FCM, and CFSFDP which showed the best performance in the previous section. The models are randomly assigned to three spots at the bottom of Fig.~\ref{fig:exp1}.A.

% %Each participant signed an informed consent form and participated under a protocol that was approved by the JHU Homewood Institutional Review Board. 

% method perf breakdown \\
% 0.864    0.870    0.812 \\ % over all 300 stimuli in behavioral experiment
% 0.860    0.772    0.753  \\% over stimuli with two clusters
% 0.868    0.831    0.852 \\ % over stimuli with three clusters
 
% Avg subject perf break down

%Perf of individual subjects will be added later to the supplement. 

%One major problem, and actually confound with the behavioral study, is that models can not automatically determine the number of clusters. [It might be possible to tweek the SFDFDP method by choosing a different set of points in the decision graph (see supplement) but that is somewhat cumbersome. Giving the number of gt clusters to fcm and SFDFDP actually gives them an advantage!! ours figure this out automatically]

\noindent \textbf{Stimuli:} We used the GMM stimuli for the user study. 
Cases where all three methods performed better than 95\% were discarded. Eventually, 300 stimuli were chosen (122 with 2 clusters and 178 with three).  

%(See appndx for sample stimuli).

%This is because of removing the ambiguity. For example, if all three did very well, then it means that the subject has to choose randomly. We wanted to avoid such cases.

%subtending a xx degree of field of view

%Stimuli were presented and responses collected using a custom script written with in MATLAB on a PC.

%Stimulus remained on the screen until the participant responded.

\noindent \textbf{Result:} Fig.~\ref{fig:user}.B shows the fraction of the cases where subjects chose models (averaged over all subjects). As it shows, subjects favored CNN output more frequently than the other two methods (p=$2.52e-06$ CNN vs. FCM; p=$0.045$ CNN vs. CFSFDP; using t-test over subjects;  n=14). This also holds over stimuli with two and three clusters. 
The average accuracies of the three clustering algorithms are shown in Fig.~\ref{fig:user}.C. As it can be seen CNN does better than other methods over the 300 stimuli.
Results of the user study indicate that our method produces clusters that are more aligned with human clustering judgments.

%Average number of selections over subjects are shown in Fig. xx. Error bars indicate xx. N

%Subjects first practiced over 5 stimuli to learn the task [this data was excluded in subsequent analysis]

% \begin{figure}[t]
% 	\centering
% 	%\footnotesize
%     \includegraphics[width=.5\linewidth]{Figs/UserStudy.pdf} \\
%     \vspace{-5pt}
% 	\caption{\small{A) A sample trial in the user study, B) Fraction of the cases where subjects chose models and break downs over clusters (all, 2, or 3), and C) Performance of the models over the stimuli.}}
% 	\label{fig:user}
%     \vspace{-15pt}
% \end{figure}

\section{Conclusion}
\label{sec:conclusion}
We argued that deep neural networks, especially CNNs, hold a great promise for data clustering. We are motivated by the fact that human vision (and learning) is a general system capable of solving both classification and clustering tasks thus blurring the current dichotomy in treating these problems. Our results show that CNNs can successfully handle complex and occluded clusters much better than other algorithms. This means that a learning mechanism, unsupervised or with minimal supervision, seems inevitable in capturing complex cluster shapes.

While our formulation is supervised, feeding the labels to the network is not always consistent. This is where our work differs from semantic segmentation and instance level segmentation. We exploited the mean squared loss to train the network. It might be possible to define other loss functions to teach the network more efficiency using less number of training data or even with weaker labels. One possibility is the pairwise accuracy that we used here for evaluation. Instead of correctly classifying labels, the emphasis can be placed on correctly predicting whether two points belong to the same cluster, regardless of cluster identities (i.e., class labels may vary).

Notice that while here we focused on synthetic stimuli, variations of the proposed CNN architecture, have been successfully applied to natural image segmentation. Thus, CNNs offer a unified solution that can be applied to different data modalities and even to higher dimensional data. Further work is needed to extend this line of work to higher dimensions, and more versatile types of cluster shapes (e.g., free form curves, Gestalt examples, density-based clusters). In this regard, adopting CNNs trained on natural images containing a rich set of intermediate- and high-level patterns can give invaluable insights.

In sum, our results provide encouragement for researchers seeking unified theoretical explanations for supervised and unsupervised categorization but raise a range of challenging theoretical questions. We emphasized more on the capacity of hierarchical frameworks and compositionality to capture complex structures rather than the learning algorithms. Indeed, further discussion and research are needed for training CNNs in unsupervised or weakly supervised manners. Some new works on learning representations from videos (e.g., for predicting future frames) are particularly interesting. Further, research on unsupervised training of spiking neural networks~\cite{masquelier2007unsupervised} and CNNs (e.g., using Hebb rule~\cite{wadhwa2016bottom}), along with computational modeling (e.g.,~\cite{serre2007robust,lecun2015deep}), and experimental studies on mechanisms of human vision (e.g.,~\cite{yamins2014performance}), will hopefully converge to computational vision algorithms that are capable of solving a variety of tasks, including classification and clustering.

\footnotesize
\bibliography{egbib}
\bibliographystyle{iclr2018_conference}
\end{document}